\pdfoutput=1

\documentclass[11pt]{article}

\usepackage{authblk}

\usepackage{ACL2023}

\usepackage{times}
\usepackage{latexsym}

\usepackage[T1]{fontenc}

\usepackage[utf8]{inputenc}

\usepackage{microtype}

\usepackage{inconsolata}

\usepackage{amsmath}
\usepackage{graphicx}
\usepackage{multirow}
\usepackage{booktabs}

\DeclareMathOperator{\sigmoid}{sigmoid}
\DeclareMathOperator{\SSF}{SSF}

\title{Lon-eå at SemEval-2023 Task 11: A Comparison of\\Activation Functions for Soft and Hard Label Prediction}

\author[1]{Peyman Hosseini}
\author[2]{Mehran Hosseini}
\author[3]{\\Sana Sabah Al-Azzawi}
\author[3]{Marcus Liwicki}
\author[1]{Ignacio Castro}
\author[1,4]{Matthew Purver}

\affil[1]{Queen Mary University of London, London, United Kingdom \authorcr
  \{\tt s.hosseini, m.purver, i.castro\}@qmul.ac.uk}
\affil[2]{King's College London, London, United Kingdom \authorcr
  \tt mehran.hosseini@kcl.ac.uk}
\affil[3]{Luleå University of Technology, Luleå, Sweden \authorcr
  \{\tt sana.al-azzawi, marcus.liwicki\}@ltu.se}
\affil[4]{Jožef Stefan Institute, Ljubljana, Slovenia \authorcr}

\begin{document}
\maketitle
\begin{abstract}
We study the influence of different activation functions in the output layer of deep neural network models for soft and hard label prediction in the learning with disagreement task. In this task, the goal is to quantify the amount of disagreement via predicting soft labels. To predict the soft labels, we use BERT-based preprocessors and encoders and vary the activation function used in the output layer, while keeping other parameters constant. The soft labels are then used for the hard label prediction. The activation functions considered are sigmoid as well as a step-function that is added to the model post-training and a sinusoidal activation function, which is introduced for the first time in this paper.
\end{abstract}

\section{Introduction}
The nuances present in natural languages, such as contextual meaning
or subjective interpretation of expressions are often discounted in
natural language processing. In subjective tasks, such as sentiment
analysis, offensive or abusive language detection, and misogyny
detection, the assumption that a gold label always exists has proven
to be an idealization \cite{Uma+21}.

Therefore, a growing body of research has been dedicated to analyzing
the disagreement from labels provided by multiple annotators. Some of
the notable works on this topic are \cite{Uma+21, Fornaciari+21,
  Uma+20, Kenyon-Dean+18}. In task 11 of SemEval 2023 on Learning with
Disagreements Le-Wi-Di, \(2^{\text{nd}}\) edition
\cite{LeWiDi2023semeval}, four datasets, including three English datasets
HS-Brexit \cite{AkhtarBP21}, ConvAbuse \cite{Curry+21}, and
MD-Agreement \cite{LeonardeLLi+21}, and one Arabic dataset, ArMIS
\cite{AlmaneaP22} have been provided. All datasets include both soft
and hard labels as well as some additional information. Among these
ArMIS, HS-Brexit, and MD-Agreement have a constant number of
annotators throughout the dataset, whereas ConvAbuse has a variable
number of annotators.

In this paper, we explore the effect of different activation functions
in the output layer of a deep neural network model for the prediction
of soft and hard labels. We introduce a sinusoidal activation
function, which we refer to as the \emph{Sinusoidal Step Function}
(\emph{SSF}). We evaluate the performance of the models using the SSF,
sigmoid, as well as a post-training step function for soft-label, and
therefore, hard-label prediction.

The experiments indicate that the SSF activation achieves the highest
F1-score for the hard-label prediction on the majority of the
datasets. For the soft label prediction, the sigmoid activation
obtains the best result for most datasets, however (cf. \autoref{sec:
  Results})\footnote{The code used for the experiments is available at
  \url{https://github.com/Speymanhs/SemEval_2023_Task_11_Lonea.git}.}.

\section{Background}
There is a dispute on how to deal with the annotator's
disagreements. Whilst some suggest that we should disregard and
discard instances with high disagreement as bad examples
\cite{Beigman-KlebanovandB+09}, others argue that such instances are
valuable and should be further analyzed and studied \cite{Uma+20,
  Fornaciari+21, Kenyon-Dean+18}. In subjective tasks, such as
offensive/abusive language detection and misogyny detection,
introducing a framework that utilizes different views and their
differences is invaluable due to the nature of such tasks, which
require integrating different viewpoints. This prompts us to utilize
this source of knowledge and try to find approaches for predicting
soft labels which will then be used to directly infer hard labels.

In the past few years transformer \cite{Vaswani+17} models, such as
BERT \cite{DevlinCLT+18} have been widely used for Natural Language
Processing (NLP) tasks. We adopt the BERT model, trained on Wikipedia
and BooksCorpus, for English datasets and the Arabic version of BERT,
also known as the AraBERT \cite{AntounFH+20}, on the Arabic ArMIS
dataset.

\subsection{Task Setup and Description}
In task 11 at SemEval 2023, we deploy approaches that use the
disagreement between annotators. By predicting soft labels, we
quantify the amount of disagreement in a range of subjective tasks
including misogyny and offensive language detection. Using the soft
labels, we also predict the hard labels that is the binary class
aligned with the majority of votes cast by the annotators for each
data.

In this approach, we use a BERT-based preprocessor and encoder in our
models to first predict soft labels. Then without any direct knowledge
or training of our models, we use the outputs computed for soft labels
and round them to the nearest integer, which represents the hard label
classes. Our analysis shows that even by using such an approach and
only training the model on soft labels, the models performs well in
predicting hard labels.



In summary, our contributions are as follows.
\begin{itemize}
\item We introduce the SSF activation for the output layer of our
  models for soft label prediction and compare its performance against
  the sigmoid activation (cf. \autoref{fig:sinAct}), as a widely used
  activation function for such tasks, and post-training step function
  (cf. \autoref{fig:stepFunc}).
\item We show that by only considering soft labels during the training
  process and without providing any prior information about the hard
  labels during the training process, the same model can be used for
  inferring hard labels.
\end{itemize}

\section{System Overview}
In the experiments, we focus on analyzing the performance of different
activation functions with the goal of finding the best performance on
the soft labels. We use three approaches for computing soft
labels. Among the four datasets of this competition, one of these
approaches, which uses the sigmoid function as activation is
applicable to all of the datasets. The other two approaches, which use
the step functions we introduce, are applicable to HS-Brexit, ArMIS,
and MD-Agreement because the number of annotators is equivalent to the
number of steps, and it is fixed in these datasets.

We start by describing our approaches for soft label prediction in
\autoref{subsec: Soft LabeL Prediction}. We then discuss our results
for predicting the hard label from the soft label in \autoref{subsec:
  Hard Label Prediction}.

\subsection{Soft Label Prediction}
\label{subsec: Soft LabeL Prediction}
In the experiments, we first choose a pre-trained BERT
model\footnote{BERT encoder available at
  \url{https://tfhub.dev/tensorflow/bert_en_uncased_L-12_H-768_A-12/4}.}.
To format the input so that the chosen BERT encoder can process it, we
first use the preprocessor\footnote{BERT preprocessor available at
  \url{https://tfhub.dev/tensorflow/bert_en_uncased_preprocess/3}.}
that transforms the input to the required format.  We concatenate the
pooled output of the BERT encoder with a dropout layer followed by a
dense layer with ReLU activation. Finally, we concatenate this dense
layer with another dropout layer, followed by a dense layer with the
activation function under the study, i.e., sigmoid, or SSF.  The
\(\sigmoid\) activation is discussed in
\autoref{subsubsection:sigApproach}, the SSF activation is discussed
in \autoref{subsubsec: SSF}, and the step function, used
post-training, is discussed in \autoref{subsubsec: Step Function}.

\subsubsection{Approach 1: Sigmoid Activation}
\label{subsubsection:sigApproach}
In this approach, we use the sigmoid function as the activation layer,
i.e., the output layer. to widen the sigmoid activation and help the
network in the learning process, we used \(\sigmoid(x/5)\) as
activation rather than \(\sigmoid(x)\). We tried a number of factors
and 5 was the one that worked best in comparison to the other factors
in the denominator. The widened sigmoid function facilitates obtaining
intermediate values between 0 and 1. The output of the sigmoid
function is then considered as the soft label.

\subsubsection{Approach 2: Sinusoidal Activation}
\label{subsubsec: SSF}
In the second approach, we introduce a sinusoidal activation function
that takes as input the number of annotators of the dataset, $a$ as
well as a slope parameter, $\theta$. The exact definition of this
function is as follows.

\begin{equation*}
\SSF_{a, \theta}(x) = \begin{cases}
             0 + \theta \cdot x & \text{if } x < 0,\\[3pt]
             f_0(x)  & \text{if } 0 \le x < \dfrac{1}{a},\\[10pt]
             f_1(x)  & \text{if } \dfrac{1}{a} \le x < \dfrac{2}{a},\\[10pt]
             \vdots & \vdots \\[10pt]
             f_i(x)  & \text{if } \dfrac{i}{a} \le x < \dfrac{i+1}{a},\\[10pt]
             \vdots & \vdots \\[10pt]

             f_{a-1}(x)  & \text{if } \dfrac{a-1}{a} \le x < 1,\\[10pt]
             1 + \theta \cdot x & \text{if } 1 \le x,\\
       \end{cases}
\end{equation*}
where $f_n(x)$ is defined as
\[f_n(x) = \dfrac{\sin(\pi\dfrac{2ax - (2n+1)}{2}) + (2n+1)}{2a}.\]

\begin{figure}
    \centering
    \includegraphics[width=\columnwidth]{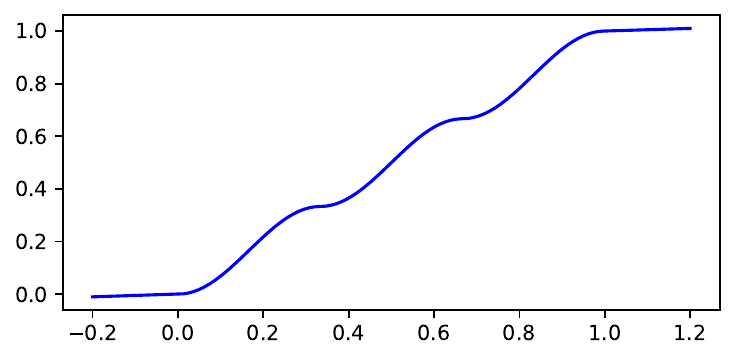}
    \caption{The plot of SSF for $\theta = 0.05$ and $a = 3$.}
    \label{fig:sinAct}
\end{figure}

\autoref{fig:sinAct} shows a plot of the \(\SSF_{a, \theta}\) for
$a=3$ and $\theta=0.05$.  The intuition behind introducing this
activation function is that the soft labels in the dataset can only
obtain certain values. For example, when the number of annotators is
3, the possible values for the soft label include 0, 0.33, 0.66, and
1.0. Therefore, an activation function that helps the prediction to
converge to one of these values has the potential to perform better in
predicting the soft labels and decreasing the loss on its prediction.

This is what our activation function tries to achieve as the slope of
the function decreases near these values and increases in between. On
the other hand, the slope of the $\sigmoid(x)$ is equal to 1 when
$x=0$ and it decreases consistently as we move x towards positive or
negative values. Even though such a slope helps the output of
$\sigmoid(x)$ to converge to 1 or 0, it is not best practice if we
want to predict intermediate values like in our case.

\subsubsection{Approach 3: Step Function}
\label{subsubsec: Step Function}
\begin{figure}
    \centering
    \includegraphics[width=\columnwidth]{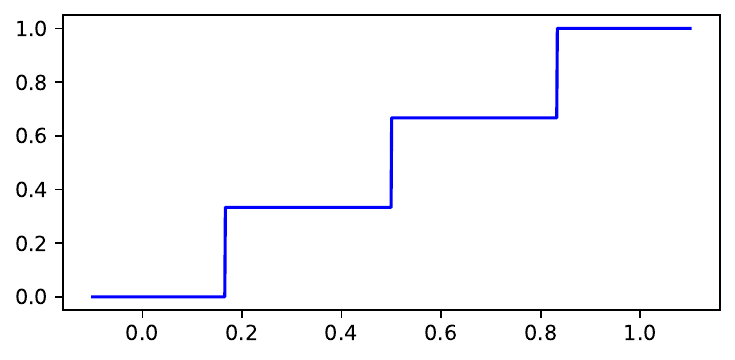}
    \caption{The discrete step function with $a = 3$.}
    \label{fig:stepFunc}
\end{figure}

In the third approach, we use a step function that takes as input the
number of annotators, $a$. This step function maps the output of the
model for the soft labels to the closest valid value. For example, in
the case of ArMIS dataset where $a = 3$, this step function is shown
in \autoref{fig:stepFunc}.

One important point to consider is that this function can't be
deployed as an activation function because its slope is 0 almost
everywhere, which prevents the model from essentially learning
anything. Therefore, the way we deploy this function in our model is
by training the model using the sigmoid activation function as
outlined in \autoref{subsubsection:sigApproach}, and then when
evaluating the results on the test set, we append this function to the
end of the model for the prediction of soft labels.

\label{section:my}
\subsection{Hard Label Prediction}
\label{subsec: Hard Label Prediction}
In addition to predicting the soft label as a metric that
quantitatively reflects the amount of disagreement on the annotators'
part, most literature still evolves around analyzing the performance
of models in the presence of gold labels and as part of this task, we
evaluate our models on hard labels as well.

The approach deployed in this experiment for deriving hard labels is
to use the same model we train for the prediction of the soft labels
without any further training and fine-tuning. We then round the
model's output for the soft label to the closest integer, which can be
either 0 or 1. Therefore, we return 0 for the hard label if the soft
label is less than or equal to 0.5, and we return 1 otherwise.

\section{Experimental Setup}
\label{sec: Experiments}
We use the official release datasets and the standard
train/test/validation splits as released by the task organizers. After
preprocessing the input using the preprocessor and passing the
preprocessed input through the pre-trained encoder, as outlined in
\autoref{subsec: Soft LabeL Prediction}, we concatenate the pooled
output of the encoder with a dropout layer with the dropout rate of
$0.2$, for regularization purposes, followed by a dense layer with 20
neurons with ReLU activation. This was then followed by another
dropout layer with a dropout rate of $0.15$. We finally, appended a
dense layer with a single neuron and one of the activation functions
discussed before, i.e., sigmoid or SSF.

During the training process, we minimize the cross-entropy loss on the
soft labels. In all approaches, we train the models for 100 epochs and
use checkpoints to save the network with the least loss on the
validation set. After training, we evaluate the performance of the
saved network on the test dataset. Finally, we round the result of
soft label predictions to the nearest integer for predicting hard
labels for the binary classification task. The metric used for
evaluating the performance of the models on the hard labels is the
micro F1-score. It is worth mentioning that evaluating performance on
hard labels using F1-score is more meaningful than a metric like
accuracy since F1-score is reflective of the model's performance on
all classes when the datasets' distribution of hard labels is skewed
towards one class, such as the ones considered in this task.

\begin{table*}[ht]
    \centering
    \begin{tabular}{lrrrr}
    \toprule
    Approach & HS-Brexit & ArMIS & ConvAbuse & MD-Agreement\\
    \midrule
    Best Result& 0.235   & 0.469 & 0.185 & 0.472\\
    Approach 1 (sigmoid) & 0.319 & 0.655 &  0.234 &  0.532\\
    Approach 2 (SSF) & 0.464 & 1.369 & - & 0.521\\
    Approach 3 (SF) &   0.710  & 2.878 & - & 0.994 \\
    Organisers Baseline & 2.715  & 8.908   & 3.484 & 7.385\\
    \bottomrule
    \end{tabular}
    \caption{\label{tab: Soft Evaluation}
    Result of soft evaluation on the test sets. the numbers in the table represent cross-entropy loss.
    }
\end{table*}
\begin{table*}[h!]
    \centering
    \begin{tabular}{lrrrr}
    \toprule
    Approach & HS-Brexit & ArMIS & ConvAbuse & MD-Agreement\\
    \midrule
    Best Result& 0.9329   & 0.8475 & 0.9493 & 0.8471\\
    Approach 1 (sigmoid) & 0.9048 & 0.7172 &  0.9310 & 0.7880\\
    Approach 2 (SSF) & 0.8929 & 0.7724 & - & 0.8037\\
    Approach 3 (SF) &   0.9048  & 0.7172 & - & 0.7880 \\
    Organisers Baseline & 0.8420  & 0.4170   & 0.7410 & 0.5340\\
    \bottomrule
    \end{tabular}
    \caption{\label{tab: Hard Evaluation}
    Result of hard evaluation on the test sets. the numbers in the table represent the micro F1-score.
    }
\end{table*}
\section{Results}
\label{sec: Results}
In this section, we discuss the results obtained on each dataset. The
focus of our experiments is to try different activation functions in
the output layer to compare the performance of these activation
functions for predicting soft labels and hard labels. Our study does
not involve analyzing and finding different splits for the data, and
we use the splits as released by the task organizers. Furthermore, we
do not train the models on additional data before training them on the
competition datasets as this adds more variables to the results of the
experiment other than the activation functions used in this
study. Also, the second and third approaches are only used for
HS-Brexit, ArMIS, and MD-Agreement datasets, as the number of
annotators in the ConvAbuse dataset is variable across different
instances. The second and third approaches are only applicable to
datasets where the number of annotators is a constant number across
the dataset instances.

We divide our discussion about the results into two parts, the results
of the soft evaluation and hard evaluation. In each subsection, we
present a table that summarises the results of the different
approaches for the corresponding metric, on all datasets. The tables
also include the results of the organizers' baseline and the best
performance achieved on each metric by the participating teams in the
competition.

\subsection{Soft Evaluation Results}
The results for the soft evaluation are shown in \autoref{tab: Soft
  Evaluation}. It is worth mentioning that the results presented for
Approaches 1, 2, and 3 in the ArMIS column are post-competition
results. The reason behind adding these post-competition results is
that in the competition, we used a different pre-trained BERT encoder
in comparison to the other three datasets. However, here, for the sake
of consistency with the approach taken for the other datasets and also
to provide a ubiquitous approach for all datasets regardless of their
language, we provide the results using the BERT Base encoder for the
Arabic language as provided in
\href{https://github.com/aub-mind/arabert}{this GitHub
  repository}. The results provided in the table for the other
datasets are the same as those submitted to the competition and use
the BERT base preprocessor and encoder we introduced in \autoref{sec:
  Experiments}.

As you can observe, in \autoref{tab: Soft Evaluation}, our first
approach that uses the sigmoid activation achieves the best result on
all the datasets except MD-Agreement. On MD-Agreement, our second
approach, which uses SSF achieves the best result in comparison to the
other approaches.

The third approach applies a step function to the outputs computed by
the sigmoid activation in approach one, which maps the computed real
values for the soft labels to the closest valid soft label. However,
as opposed to what we initially expected, this approach does not help
in obtaining lower losses and performs poorly in comparison to other
approaches by a large margin. We think this poor performance is
potentially due to the fact that this is a function that is applied
outside the training procedure and therefore is not optimized to
minimize the loss function.

\subsection{Hard Evaluation Results}
The results for hard label evaluation are presented in \autoref{tab:
  Hard Evaluation}. Similar to the previous subsection, the results
written for Approaches 1, 2, and 3 in the ArMIS column are
post-competition results. The way we derive the hard labels is to
directly use the output of the corresponding approach for soft label
prediction and round the value to the nearest integer. As the results
show in \autoref{tab: Hard Evaluation}, on MD-Agreement and ArMIS
dataset, the approach that uses SSF activation achieves the best
result. This difference is especially significant in the ArMIS dataset
where the SSF activation archives micro F1-score that is 5.52\% higher
than the other approaches.


\section{Conclusion}
We studied the effect of various activation functions in the output
layer of a deep neural network model, including the sinusoidal
activation function SSF, introduced for the first time in this paper,
on soft and hard label prediction in the learning with disagreement
task at SemEval 2023.

The sinusoidal activation function SSF, which can be applied to any
domain or dataset where the number of annotators is constant or their
views are summarised into a constant number of categories, shows
promising results in the hard evaluation. On two of the three datasets
that SSF was applied to, i.e., ArMIS and MD-Agreement), the use of SSF
improves the micro F1-score by about \(5.52\%\) and \(1.57\%\),
respectively, in comparison to the widely used \(\sigmoid\)
activation.

Based on the promising performance of the SSF activation for the hard
label prediction, we plan to further study the use of SSF activation
on other datasets and tasks, where a constant number or class of
annotators have annotated the data.

\section*{Acknowledgements}
This work is partially supported by the UK EPSRC via the Centre for
Doctoral Training in Intelligent Games and Game Intelligence (IGGI;
EP/S022325/1) and the projects Sodestream (EP/S033564/1), AP4L
(EP/W032473/1), REPHRAIN (EP/V011189/1) and ARCIDUCA (EP/W001632/1);
as well as the Slovenian Research Agency via research core funding for
the programme Knowledge Technologies (P2-0103) and the project SOVRAG
(Hate speech in contemporary conceptualizations of nationalism,
racism, gender and migration, J5-3102). We also thank Zahraa Al Sahili
for providing insightful comments during the early stages of this
work.

\bibliography{SemEval_Task}
\bibliographystyle{acl_natbib}

\end{document}